\documentclass[a4paper,conference]{IEEEtran}
\usepackage{times}

\usepackage{soul}
\usepackage{url}
\usepackage[hidelinks]{hyperref}
\usepackage[utf8]{inputenc}
\usepackage[small]{caption}
\usepackage{graphicx}
\usepackage{amsmath}
\usepackage{booktabs}
\usepackage{graphicx}
\usepackage{amsmath,amssymb} 
\usepackage{color}
\usepackage{todonotes}

\usepackage{algorithm}
\usepackage{algpseudocode}  
\usepackage{multirow}
\usepackage{booktabs}
\usepackage{subfigure}
\usepackage{hyperref}
\usepackage{color,soul}
\usepackage{amsmath}
\usepackage{hhline}
\usepackage{multicol}
\usepackage[normalem]{ulem}

\ifCLASSINFOpdf
\else
\fi
\usepackage{marvosym}

\begin{document}
%
\title{Neural Architecture Searching for Facial
Attributes-based Depression Recognition}

\author{\IEEEauthorblockN{Mingzhe Chen\textsuperscript{1}, Xi Xiao\textsuperscript{1}, Bin Zhang\textsuperscript{2,\Letter}, Xinyu Liu\textsuperscript{3}, Runiu Lu\textsuperscript{1}}
\IEEEauthorblockA{\textsuperscript{1}Tsinghua Shenzhen International Graduate School, Tsinghua University, Shenzhen, China\\
\textsuperscript{2}Department of New Networks, Peng Cheng National Laboratory, Shenzhen, China \\
\textsuperscript{3}Shenzhen Jinzhou Precision Technology Corp., Shenzhen China \\
cmz19@mails.tsinghua.edu.cn,
\{xiaox, lurn\}@sz.tsinghua.edu.cn, \\
bin.zhang@pcl.ac.cn,
liuxinyu@chinadrill.com}}


%


\maketitle

\begin{abstract}

Recent studies show that depression can be partially reflected from human facial attributes. Since facial attributes have various data structure and carry different information, existing approaches fail to specifically consider the optimal way to extract depression-related features from each of them, as well as investigates the best fusion strategy. In this paper, we propose to extend Neural Architecture Search (NAS) technique for designing an optimal model for multiple facial attributes-based depression recognition, which can be efficiently and robustly implemented in a small dataset. Our approach first conducts a warmer up step to the feature extractor of each facial attribute, aiming to largely reduce the search space and providing customized architecture, where each feature extractor can be either a Convolution Neural Networks (CNN) or Graph Neural Networks (GNN). Then, we conduct an end-to-end architecture search for all feature extractors and the fusion network, allowing the complementary depression cues to be optimally combined with less redundancy. The experimental results on AVEC 2016 dataset show that the model explored by our approach achieves breakthrough performance with 27\% and 30\% RMSE and MAE improvements over the existing state-of-the-art. In light of these findings, this paper provides solid evidences and a strong baseline for applying NAS to time-series data-based mental health analysis. 

\end{abstract}

\IEEEpeerreviewmaketitle

\section{Introduction}

\noindent Major depressive disorder (MDD) is the most common mental illness with around $5.07\%$ of adults in the world suffering from it \cite{GHDx}. 
While the standard clinical depression assessment relies on structured interviews by specially trained psychiatrists \cite{first2004structured}, it is subjective, time-consuming and hard to access. 
There is convergent evidence suggests that non-verbal facial behaviors provide rich and reliable sources for reflecting human depression status \cite{jaeger1986facial,gaebel2004facial} (e.g., depressed patients usually have reduced facial expressions), and they are easily to be recorded in non-invasive ways using portable devices (mobile phone, laptop, etc.). 
As a result, a large number of recent studies attempt to automatically recognize depression status from subjects' faces.


Standard face-based approaches \cite{zhou2018visually,he2021automatic,de2020encoding,de2021mdn} predict depression directly from face images/videos. However, in real-world applications, face images are sometimes not accessible due to various ethical and privacy policies. Since an early study \cite{cohn09} show that mid and low-level facial attributes (e.g., Facial Action Units (AUs) and facial landmarks) are informative for depression status, a certain number of recent studies devote to recognize depression from automatic detected facial attributes such as facial landmarks \cite{yang2016decision,Nasir2016,Syed2017}, gaze direction \cite{alghowinem2015cross,alghowinem2016multimodal}, facial action units (AUs) \cite{gong2017topic,Song2018,Song2020}, and head poses \cite{yang2016decision}. Besides some of them compute several statistics \cite{jaiswal2019automatic,yang21} (e.g., displacement, velocity, acceleration) from facial attributes time-series as the clip-level representation for depression recognition, recent advances in deep learning (e.g., 1D-CNN \cite{Song2018,Song2020}, LSTM \cite{Ray2019}, attention-based temporal CNN \cite{Du2019}, Causal CNN \cite{Haque2018}, etc.) also have been applied to infer depression from facial attribute time-series, and achieved enhanced results over most hand-crafted approaches.

However, all of these approaches only manually design models (hand-crafted feature extraction or manually designed CNNs) to extract features from facial attributes, and conduct simple fusion strategy to combine depression cues extracted from all attributes, e.g., they conduct standard decision-level fusion or simply concatenate all facial attributes as a joint representation. Since each facial attribute has a unique data structure, existing approaches fail to design a task-specific architecture for each facial attribute's feature extraction. Moreover, while each facial attribute contains both unique and common cues (also carried by other facial attribute) for depression recognition, these simple fusion strategies can not optimally retain complementary cues and minimize the redundancy from all facial attributes. In other words, existing approaches that manually design networks are not able to optimally extract and combine depression-related features from multiple facial attributes, which would theoretically limit the recognition performance.

\begin{figure*}[ht]
    \centering
    \includegraphics[width=0.9\textwidth]{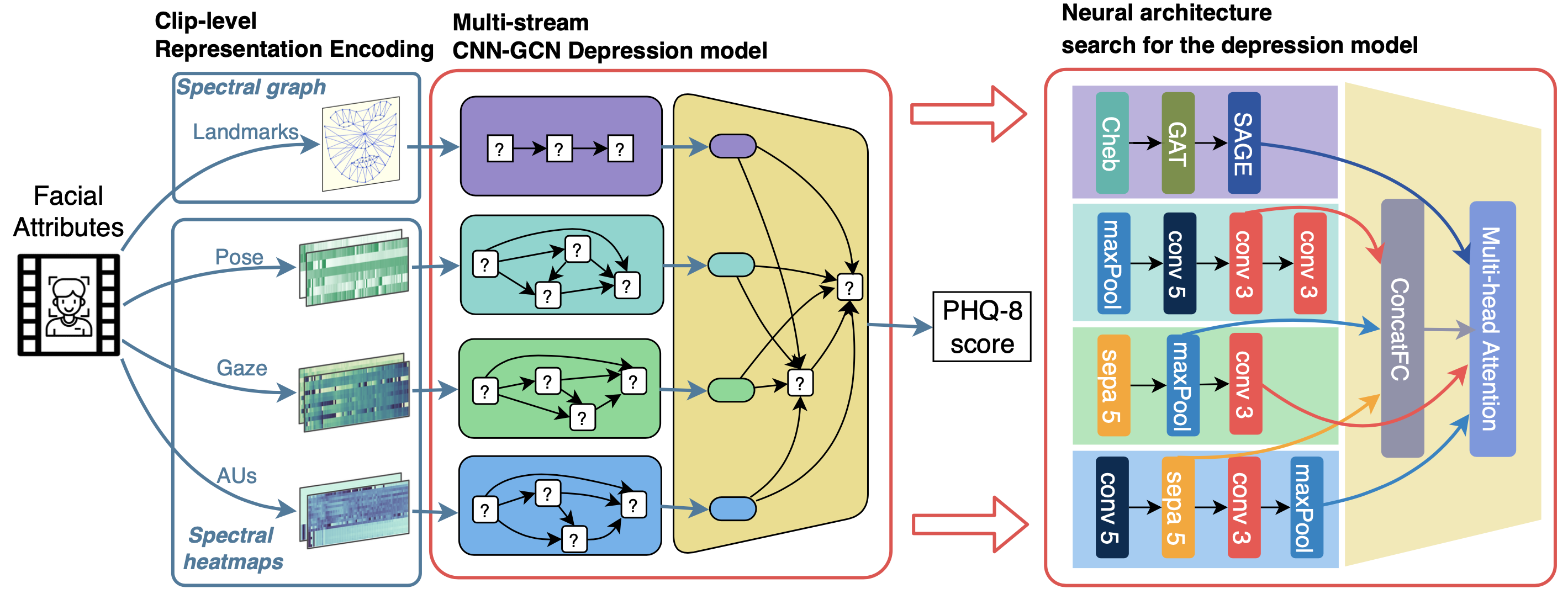}
    \caption{Illustration of the proposed approach. We first encode facial attribute time-series of the clip to a pair of fixed-size clip-level spectral representations (Sec. \ref{sec:clip-level}). Then, we propose a multi-stream model to process multiple clip representations (Sec. \ref{sec:network}). Finally, we conduct end-to-end Neural Architecture Search to obtain an optimal model, where each stream is customized to learn depression-related cues from the unique facial attribute while the fusion module making the best combination of their latent representations (Sec. \ref{subsec: NAS}).}
    \label{fig:pipeline}
\end{figure*}

In this paper, we address the aforementioned issues by introducing Neural Architecture Search (NAS) technique to explore an optimal model from small depression dataset, for multiple facial attributes-based depression recognition. Instead of conducting frame/short segment-level depression modelling, our approach starts with employing the spectral encoding method \cite{Song2020} to obtain a clip-level facial behavioral representation for each subject, which has been frequently claimed
duo be more reliable for depression recognition. Then, we propose a novel multi-stream CNN-GCN framework, where each stream is specifically customized to the unique data structure of a specific facial attribute, aiming to learn extract depression-related features from its clip-level representation, while its fusion module selects a best intermediate latent representation from each stream and conducting optimal operations for all representations' fusion. To achieve this optimal network, we propose an end-to-end NAS strategy to jointly search task-specific architectures for all modules with limited number of depression data. The proposed multi-stream CNN-GCN framework is illustrated in Fig. \ref{fig:pipeline}. The main contributions and novelties of our approach is summarized as follows.

\begin{itemize}
    
    \item We propose a novel multi-stream framework for multiple facial attributes-based automatic depression recognition, where each stream can be either a CNN or a GNN, and the fusion module optimally combines most informative latent representations that are produced by these streams for depression recognition. To the best of our knowledge, this is the first CNN-GCN framework for face-based depression recognition.

    \item We propose a Neural Architecture Search (NAS) method to search an optimal multi-stream CNN-GCN framework from a depression dataset that only contains 107 training samples, where a novel motion average loss function is proposed to stabilize the searching process. To the best of our knowledge, this is the first work that extends the Neural Architecture Search (NAS) technique to automatic depression analysis. 
    
    \item The experimental results show that our approach achieved new state-of-the-art results with 27\% RMSE and 30\% MAE improvements over the previous state-of-the-art method \cite{Du2019,yang21}.
    
\end{itemize}

\section{Related work}

\subsection{Facial attributes-based depression recognition}


Due to the ethical concern or storage issue, recent depression recognition challenges \cite{Valstar2016,Ringeval2017,ringeval2019avec} encourage researchers to recognize depression from automatically detected facial attributes (e.g., AUs, emotions, facial landmarks, etc.). Jaiswal et al. \cite{jaiswal2019automatic} summarize video-level facial attribute time-series into a histogram, and feed it to MLP to predict the target person's depression level. Yang et al. \cite{yang21} introduce a novel hand-crafted video descriptor that manually models the dynamics of 2D facial landmarks of each video segment, which is then fed to a CNN for segment-level depression-related feature extraction. Haque et al. \cite{Haque2018} use a Causal Convolutional Neural Network (C-CNN) to deep learn sentence-level depression cues from 3D facial landmarks. Du et al.\cite{Du2019} propose a Atrous Residual Temporal Convolutional Network (DepArt-Net) that generates multi-scale contextual features from several low-level visual behaviors, then temporally fuse them through attention mechanism to capture the long-range depression-related cues. Song et al.\cite{Song2018,Song2020} propose to use Fourier transforms to encode facial attribute time-series (AUs, gazes, and head poses) of a clip into a length-independent spectral representation, incorporating multi-scale temporal information. However, all these approaches manually design networks for depression feature extraction and fusion without considering the task-specific architecture.

\subsection{Neural Architecture search}

\noindent Neural Architecture Search (NAS) is an AutoML technique that allows to automatic design a task-specific artificial neural network architecture for the target. Early pioneering work \cite{Zoph2017} treats neural network architecture search as a combinatorial optimization problem on a discrete search space. A controller optimized by reinforcement learning (RL) \cite{Zoph2017,baker2016designing, zhong2017practical,Pham2018,tan2019mnasnet} or evolutionary algorithms \cite{xie2017genetic,real2019regularized,wei2020npenas} iteratively propose plausible network architectures, where validation results obtained by the explored network is used as a reward signal to update the controller, enforcing it to propose a better architecture. While such strategies achieved promising performance, they are time-consuming. To accelerate searching process, ENAS \cite{Pham2018} shares network's parameters for all candidate architectures. It treats the entire search space as a super computational graph, and the candidate neural architecture can be viewed as a directed acyclic subgraph. By sharing the model weights among all the different subgraphs (candidate architectures), it is possible to avoid training each subgraph completely from scratch. Recently, Liu et al. \cite{Liu2019} propose a DARTs model that replaces the discrete searching process with a continuously differentiable strategy, allowing gradient descent-based architecture optimization and resulting in exponentially faster searching speed. However, such continuously differentiable strategy is less likely to explore an optimal architecture \cite{yu2019evaluating,zhang2020overcoming,chu2021fairnas,li2020improving} compared with RL-based methods.

\begin{figure}[ht]
    \centering
    \includegraphics[width=0.5\textwidth]{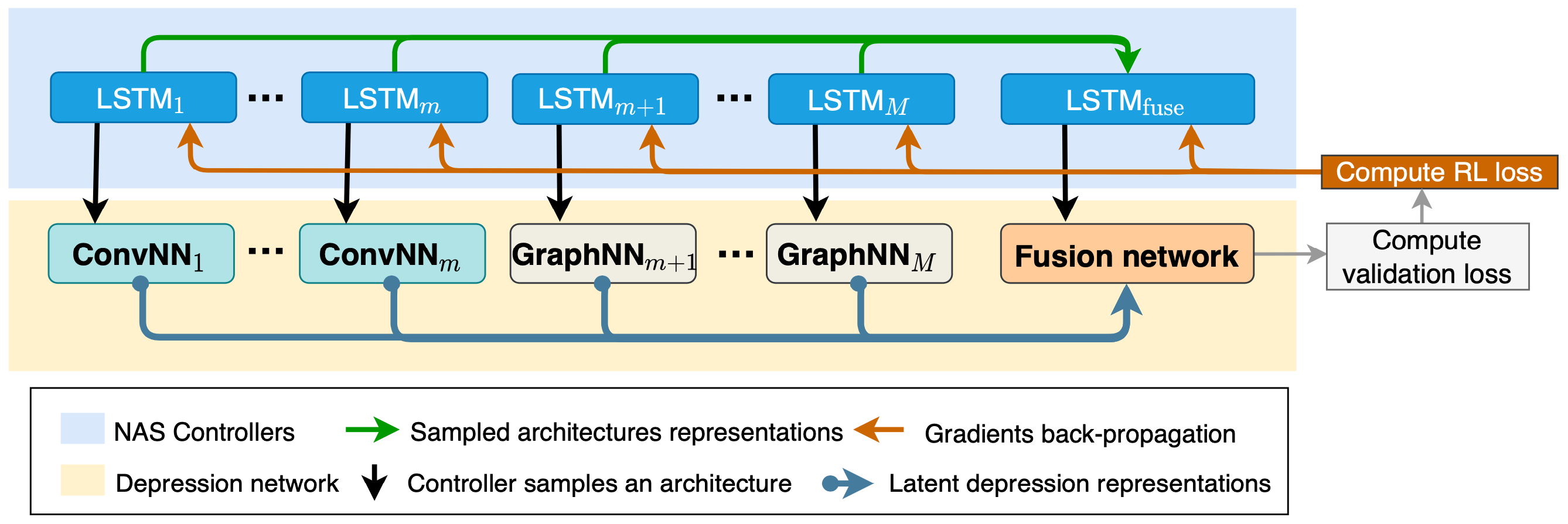}
    \caption{Illustration of our end-to-end NAS strategy. The loss values between the predictions generated by the fusion module and the ground-truth back-propagate to all controllers (LSTMs), enforcing them to jointly sample better architectures for all streams and the fusion module that make better depression prediction.}
    \label{fig:nas}
\end{figure}

\section{Methodology}


\noindent In this section, we present the details of our depression recognition approach. Our approach first converts all facial attribute time-series of each clip to spectral representations, which provides a set of length-independent clip-level facial attribute representations, summarising various long-term facial behaviors of the target person (Sec. \ref{sec:clip-level}). Then, we describe the proposed multi-stream CNN-GCN depression model in Sec. \ref{sec:network}, which aims to learn task-specific and complementary depression cues from the clip-level representation of each facial attribute while optimally combining them for depression recognition. Finally, we propose an end-to-end strategy to search for the target depression model from a small depression dataset (Sec. \ref{sec:clip-level}).


\subsection{Clip-level facial attribute representations encoding}
\label{sec:clip-level}

\noindent As discussed above, depression status is more reliable to be reflected from long-term facial behaviors. In this sense, the first step of our approach is to summarize a clip-level representation for the target clip. Since the length of recorded face clip can be various for different subjects, the facial attribute time-series of each subject would be variable. To this end, we apply the spectral encoding algorithm \cite{Song2020} to produce a pair of fixed-size spectral representations from facial attribute time-series of each clip, which summarize multi-scale facial behavioral dynamics. Specifically, given $N$ facial attributes $f_m$ ($m = 1, 2, \cdots, M$) of a clip with an arbitrary length, each of them is a time-series with $C_n$ channels. Then, their spectral 
representations can be denoted as $S_m^{\text{amp}} \in \mathbb{R}^{C_m \times K}$ and $S_n^{\text{pha}} \in \mathbb{R}^{C_m \times K}$, where $S_n^{\text{amp}}$ and $S_n^{\text{pha}}$ are amplitude and phase spectra of the $m_{th}$ facial attribute; $K$ is a pre-defined hyperparameter of retained frequency components (the number of columns) for the spectra. As a result, facial attributes of clips with various lengths can be summarized into clip-level spectral representations with the same size.

\subsection{Multi-stream CNN-GNN depression analysis model}
\label{sec:network}

\noindent As illustrated in Fig. \ref{fig:pipeline}, the proposed model consists of multiple feature extractors and a fusion module. Each feature extractor aims to extract depression-related cues from the produced spectral representations of a specific facial attribute. In particular, we define a feature extractor as either a CNN or a GNN depending on the data structure of the facial attribute. We first categorize facial attributes into two types: (1) facial attributes whose channels do not have clear spatial correlations (e.g., AUs, gaze and head pose); and (2) facial landmarks that have strict spatial distributions. For each facial attribute time-series of the first type, we concatenate its clip-level spectral representations as a multi-channel heatmap $S^n = [S_n^{\text{amp}}; S_n^{\text{pha}}] \in \mathbb{R}^{2 C_n \times K}$, and use a 1D-CNN to deep learn depression-related feature from it. For facial landmark time-series, we represent them as a clip-level graph. Specifically, we concatenate spectral representations of each facial landmark as a single spectral vector, which is represented as a node feature in the graph. Here, nodes that belongs to a certain facial region (e.g., eyes or mouth) are fully connected, and all nodes are connected to the node that describes the nasal root, allowing messages can be exchanged for different facial regions during the GNN processing.

Meanwhile, we also propose a fusion module to optimally combine depression features produced by all feature extractors. It contains a input block and several fusion blocks. The input block consists of $N$ parallel input layers, each of which takes the best latent feature from a specific feature extractor. Here, we use the average pooling that aligns all latent representations to the same size, which are then fed to several fusion blocks to optimally combine all depression cues. The employed candidate operators for each feature extractor and the fusion module are provided in the supplementary material.

\subsection{Neural Architecture Search for the multi-stream model}
\label{subsec: NAS}

\noindent To obtain an optimal multi-stream model to predict depression from multiple facial attributes, this sections proposes to automatically search for the optimal architecture of the model under the supervision of depression labels. Our searching strategy is made up of two stages: the warm-up stage and the end-to-end depression model searching stage.



\subsubsection{Warm-up stage} Let's assume $T_m$ is the search space size for the $m_{th}$ feature extractor, and $T_f$ is the search space size for the fusion module. If we directly conduct the end-to-end searching for all feature extractors and the fusion module, the search space size for the entire model can be denoted as 
\begin{equation}
C = T_f \times \prod_{m = 1}^M T_m  
\end{equation}
The $C$ is intimating when the search space size $T_m$ for each module is large. To this end, the warm-up stage first conducts a pre-searching for each feature extractor to reduce the search space size from $T_m$ to $\hat{T}_m \ll T_m$. Consequently, the searching complexity of the warm-up stage is $\sum_{m = 1}^M \times T_m$, while the complexity of end-to-end searching is reduced to $T_f \times \prod_{m = 1}^M \hat{T}_m$. In short, the warm-up stage reduces the depression network searching complexity to 
\begin{equation}
C = \sum_{m = 1}^M T_m + T_f \times \prod_{m = 1}^M \times \hat{T}_m
\end{equation}
For each feature extractor, we learn a Long-short-term-memory network (LSTM) as the controller to sample architectures. The pseudocode of the warm-up stage is demonstrated in the supplementary material.

\begin{algorithm*}
\algrenewcommand\algorithmicrequire{\textbf{Input:}}
\algrenewcommand\algorithmicensure{\textbf{Output:}}
\caption{Joint Neural Architecture Search for our multi-stream depression model}\label{algo:joint}
\begin{algorithmic}[1]
\Require Unified search space $\bigcup_m \{\mathcal A^m\}$, Training and Validation datasets $\mathcal D_{train}, \mathcal D_{val}$, single-modal controllers $\{\pi_{\theta^m}\}$, fusion controller $\pi_{\theta^f}$, Monte-Carlo steps $N$, Motion average tracker $\mathbb D$, Max timesteps $T$
\Ensure Optimal multi-modal fusion network for depression recognition
\State initialize fusion controller parameters $\pi_{\theta^f}$
\State initialize or load warm-up parameters for single-modal controllers $\{\pi_{\theta^m}\}$
\State initialize motion average tracker $\mathbb D\leftarrow \{\}$ 
\State denote parameters of all controllers as $\theta = \big(\{\theta^m\}, \theta^f\big)$  

\For{$t=1$ \textbf{to} $T$}
	\State $L(\theta, \theta_t) \leftarrow 0$ 
    \For{$i=1$ \textbf{to} $N$}
        \For{\text{each modality} $m$}
            \State $\pi_{\theta^m}$ samples a single-modal architecture $\mathbf A_i^m \sim \mathcal A^m$ and output embedding $e_i^m$
            \State instantiate single-modal networks $\mathcal M_i^m$ based on $\mathbf A_i^m$
        \EndFor
        \State $\pi_{\theta^f}$ samples the fusion architecture $\mathbf A_i^f \sim \mathcal A^{fuse}$ based on $\{e_i^m\}$
        \State instantiate the fusion network $\mathcal M_i^f$ based on $\mathbf A_i^f$
        \State train the joint model $\left(\{\mathcal M_i^m\}, \mathcal M_i^f\right)$ on $\mathcal D_{train}$ from scratch
        \State evaluate $\left(\{\mathcal M_i^m\}, \mathcal M_i^f\right)$ on $\mathcal D_{val}$ to get validation error $\mathcal E_{val}$ 
        \State update $\mathbb D$ with $(\mathbf A_i, \mathcal E_{val})$
        \State compute and accumulate $L(\theta, \theta_t)$ with $\mathbb D[a_i]$
        \EndFor
	\State update $\theta$ with $\nabla L(\theta, \theta_t)$ \Comment{See formula(\ref{eq:2})}
\EndFor
\State \Return $B$
\end{algorithmic}
\label{alg:end-to-end}
\end{algorithm*}

\subsubsection{End-to-end architecture searching stage} Once we obtained the reduced search space for each feature extractor, we then jointly search for the final architecture of all feature extractors and the fusion module. To achieve a high depression recognition performance, this joint searching process enforces all feature extractors to be explored to learn unique and complementary depression cues rather than repeatedly learn depression cues that contained in all facial attributes. Specifically, we individually learn a LSTM as the controller to sample architectures for each feature extractor, all of which are jointly optimized with the controller that samples the architecture for the fusion module. This process is demonstrated in Algorithm \ref{alg:end-to-end} and Fig. \ref{fig:nas}.

\subsubsection{Optimization details} At each training iteration $t$, the controller (LSTM) samples an architecture $\mathbf A$ based on its current policy $\pi_{\theta_t}$. Then, we instantiate a child network $\mathcal M$ from $\mathbf A$ and train it from scratch. The performance of the trained child network is evaluated by the Root Mean Square Error (RMSE) between its predictions and the ground-truth on the valiation set, which is used as the reward signal $r(\mathbf A)$ to train the controller via the PPO algorithm\cite{schulman2017proximal}. Let $\theta_t$ denote the parameters of the controller at the time step $t$ and $\pi_{\theta}(\mathbf A)$ denote the probability of $\mathbf A$ under policy $\pi_{\theta}$. The PPO objective for sampled architecture $\mathbf A$ is formulated as:
 \begin{align} \label{eq:1}
 \begin{split}
  J_{ppo}\left(\mathbf A, \theta_t, \theta\right)&=\min \Big(\frac{\pi_{\theta}(\mathbf A )}{\pi_{\theta_t}(\mathbf A )}  r(\mathbf A), g\left(\epsilon, r^{\pi_{\theta_{t}}}(\mathbf A)\right)\Big)
 \end{split}
 \end{align}
 where
 \begin{equation*}
 \quad g(\epsilon, r)= \begin{cases}(1+\epsilon) r, r \geq 0 \\ (1-\epsilon) r, r<0\end{cases}
 \end{equation*}
 During the joint search, the joint probability for the fusion architecture can be factorized as
  \begin{equation}
\pi(\mathbf A) = \pi(\mathbf A^f; \{\mathbf A^m\}) = \pi_{\theta_f}(\mathbf A^f|\{\mathbf A^m\})\prod_{m=1}^M \pi_{\theta_m}(\mathbf A^m) 
\end{equation}
where $\mathbf A^f$ and $\{\mathbf A^m\}$ stands for the architectures of the fusion module and each feature extractor.
Finally, the loss function for updating the controller is approximated by an average:
 \begin{align} \label{eq:2}
 \begin{split}
 L(\theta,\theta_t)=\mathbb E_{\mathbf A \sim \pi_{\theta_t}}[J_{ppo}\left(\mathbf A, \theta_t, \theta\right)] \approx \frac{1}{N}\sum_{i=1}^N J_{ppo}\left(\mathbf A_i, \theta_t, \theta\right)
 \end{split}
 \end{align}
where $\{\mathbf A_i\}_{i=1}^N$ are the architectures sampled from $\pi_{\theta_t}$.

Since the size of the depression recognition dataset is very limited, the trained models of a certain sampled architecture $\mathbf A$ can have variable performances due to the stochastic factors of the training (e.g., the shuffled training samples or different initial weights). This means that the validation error in a single trial can not accurately represent the generalization capability of the sampled architecture. 
In principle, we can mitigate this problem using the Monte Carlo method, averaging the performance by training and evaluating $\mathbf A$ multiple times each time $\mathbf A$ is sampled. 
However, the number of trials used for averaging, denoted as $N$, is difficult to determine, and a large $N$ entails significant computational costs. 
In this sense, we modify the Monte Carlo method with a motion average (line 17 of the Algorithm \ref{algo:joint}). This process replaces the validation error, i.e. the reward signal, in a single trial with the average results achieved across the whole training process for $\mathbf A$.  The proposed motion average is formulated as
\begin{equation}
\overline{r}(\mathbf A) =\overline{\mathcal E_{val}^{\mathbf A}} = \frac{1}{N}\sum_{i=1}^N \mathcal E_{val}^{\mathbf A, i}
\end{equation}
where $\mathcal E_{val}^{\mathbf A, i}$ is the validation loss obtained by the $i_{th}$ instantiated child network of the architecture $\mathbf A$.
Intuitively, the more frequently an architecture is sampled, the more accurate its performance estimate is, allowing to obtain a more accurate ranking of architectures when the policy converges in the later stage.

\section{Experiment}
\label{Experiment}


\subsection{Dataset}
\label{dataset}

\noindent In this paper, we evaluate the proposed approach on a widely-used facial attributes-based depression analysis dataset: DAIC-WOZ dataset\cite{gratch2014distress}. The DAIC-WOZ dataset used in the AVEC 2016 depression recognition challenges contains a total of 189 clinical interviews\cite{Nasir2016}.
The interviews are conducted by an animated virtual interviewer called Ellie, ranging from 7 to 33 minutes in length. 
Each interview session contains multiple audio-visual and verbal recordings of the participants answering Ellie's questions, as well as self-assessed PHQ-8 scores(0-24) as ground truth labels.


\subsection{Implementation details}
\label{details}

\noindent \textbf{Pre-processing:} For each clip, we first remove frames whose face detection are failed or have low confidence of the detected face. We then normalize each of its facial attribute time-series by subtracting the median of the time-series (we retain original values for facial landmark time-series to retain spatial information). Meanwhile, during the spectral encoding, we retained $K=120$ frequency components of the spectral signals for all features, following the frequency alignment method in \cite{Song2020}.

\textbf{Training details:}
\label{setting}
For CNN networks and fusion blocks, in addition to the stem architecture sampled by the controller, a fixed regression layer, i.e., average pooling+dropout+linear, is used to output the final prediction in the warm-up stage.
For GNN networks, only dropout+linear is added, since the pooling layer is included in the NAS search space.The loss functions of the controller and child networks are the same in both warm-up and fusion stages. During the NAS for fusion, the cell layers of the single-modal controller LSTMs are concatenated as the cell and the input of the first layer of the fusion controller LSTM. 
The controller and child networks are both optimized with Adam.
The hyperparameters of the controller and child networks are optimized on a validation set and keeps same in all experiments.
Finally, to obtain the best architecture, we trained the top-3 architectures proposed by the controller on the full training set and report the best performing architecture on the test set.

\textbf{Metrics:} We adopt the two metrics (i.e., root mean square error(RMSE) and mean absolute error(MAE)) to evaluate the performance of our approach, which have been used in previous AVEC challenges. They are defined as:
\begin{equation}
 \text{RMSE} =\sqrt{\frac{1}{N} \sum_{i=1}^{N}\left(\hat{y}_{i}-y_{i}\right)^{2}},
\end{equation}
\begin{equation}
\text{MAE}=\frac{1}{N} \sum_{i=1}^{N}\left|\hat{y}_{i}-y_{i}\right|
\end{equation}
where $\hat{y}_i$ and $y_i$ denote predictions and the ground truth PHQ-8 depression scores.


\subsection{Comparison to existing approaches}\label{sota}

\begin{figure}[ht]
    \centering
    \includegraphics[width=0.35\textwidth]{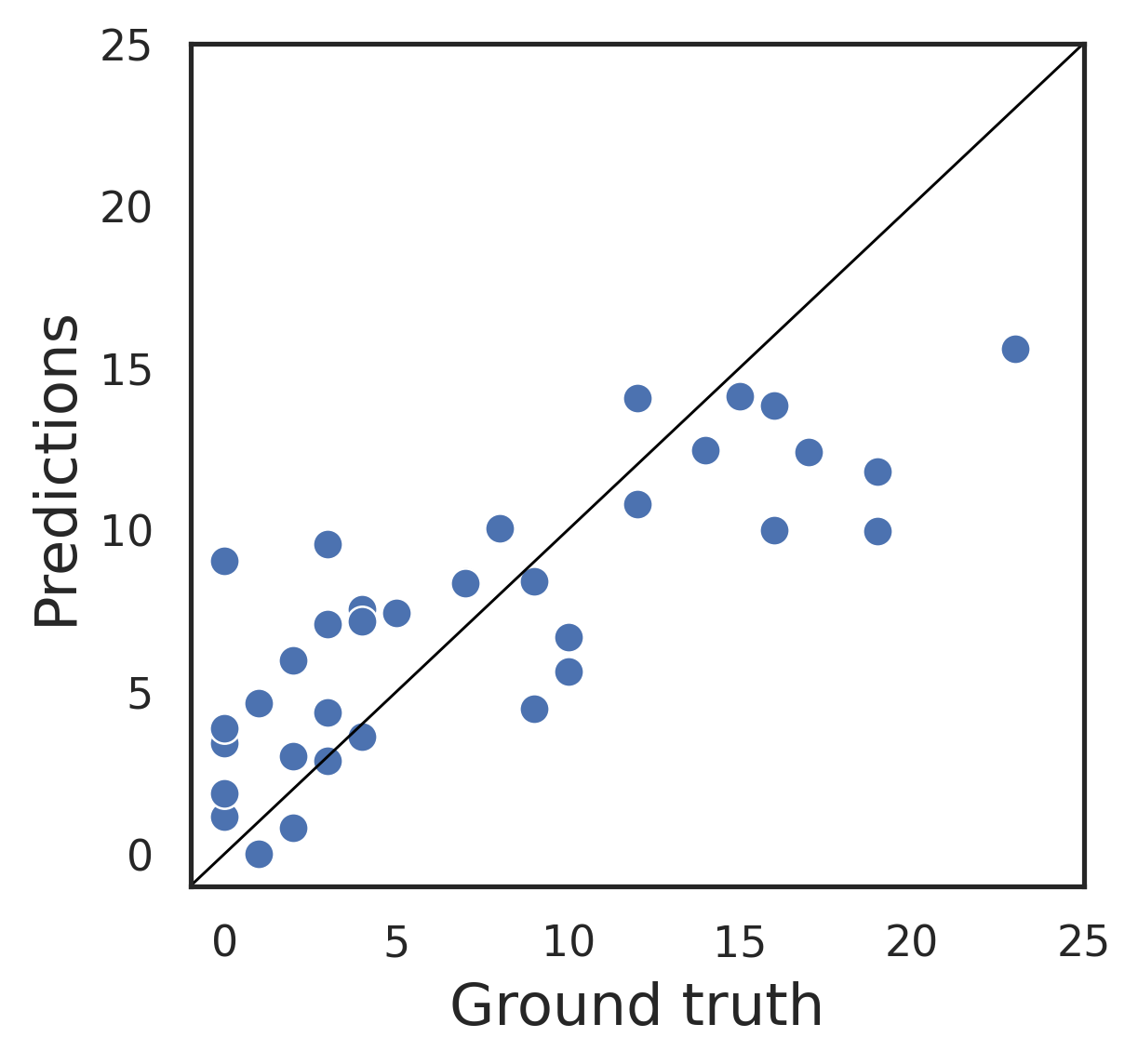}
    \caption{Predictions of the optimal architecture}
    \label{fig:3}
\end{figure}

Table \ref{table:1} compares our best system to recent state-of-the-art methods. It is clear that our method achieves the new state-of-the-art performance with ~27\% RMSE and ~30\% MAE improvement over the existing state-of-the-art facial attributes-based approach \cite{yang21}. A more detailed analysis is shown in Table \ref{table:2}, where we report the depression recognition results achieved by each explored single-modal network (each predict depression from a single facial attribute). The results demonstrate that the automatically searched architecture provide promising results for all attributes, with large advantages over other methods. Particularly, our approach obtained the top-2 best performance from the GNN-based facial landmark stream and CNN-based AU stream, which show that even using a single modality achieves better performance than any existing facial attribute-based approach. In other words, the aforementioned results suggest that our approach allows to search superior network architecture than existing manually designed architectures for extracting depression cues from each facial attribute, i.e., existing approaches can not explicitly extract depression cues from each facial attribute. More importantly, these results indicates that there is a great potential of applying NAS for automatic depression analysis. Fig. \ref{fig:3} visualizes the predictions of our best system.

\begin{table}[t]\centering
   \begin{tabular}{ p{3cm}  c c}
    \hline
    \hline
     Method  & RMSE & MAE \\
        \hline
    Baseline & 7.13 & 5.88 \\
    Williamson et al. \cite{Williamson2016} & 6.45 & 5.33 \\   
    Song et al. \cite{Song2018}   & 6.29  & 5.15 \\  
    Haque et al. \cite{Haque2018}                  & - & 5.01 \\ 
    Du et al.  \cite{Du2019}       & 5.78 & 4.61 \\  
    Yang et al. \cite{yang21}      & 5.39& 4.72 \\  
    Ours           & \textbf{3.96}& \textbf{3.23} \\
    \hline
    \hline
   \end{tabular}
   \caption{Comparison between our approach to others on AVEC 2016 dataset.}
   \label{table:1}
\end{table}

 \begin{table}\centering
    \begin{tabular}{p{2cm} ccccc}
     \hline
     \hline
     \quad Method & AUs  & Gaze & Pose & Landmarks & Fusion  \\
      \hline
     \vspace{0.05\baselineskip}
         \textit{RMSE}                       \\
        \quad                    Song et al. \cite{Song2018}  & 6.32  & 6.36  & 6.18 &  -   & 6.29  \\
             \quad                Du et al. \cite{Du2019}    & 5.88  & 6.21  & 6.32 & 6.02 & 5.78  \\
                  \quad           Ours          & 5.11  & 5.94  & 5.61 & 4.93 & 3.96  \\
     \hline
     \vspace{0.05\baselineskip}
    \textit{MAE}    \\
    \quad  Song et al. \cite{Song2018}  & 5.01  & 5.24  & 5.04 &  -   & 5.15  \\
         \quad                     Du et al.  \cite{Du2019}   & 4.65  & 4.99  & 5.21 & 5.01 & 4.61  \\
        \quad                      Ours          & \textbf{4.38}  & \textbf{4.83}  & \textbf{4.62} & \textbf{3.82} & \textbf{3.23}  \\
     \hline
     \hline

    \end{tabular}
    \caption{Results achieved by each explored facial extractor and the fusion module.}
\label{table:2}
 \end{table}

\subsection{Ablation studies} 
\label{ablation}

\noindent According to Table \ref{table:2}, we first notice that facial landmark time-series contain more depression-related cues than other facial attributes. This can be explained by the fact that facial landmarks can comprehensively describe behaviors of various facial regions. Although AUs can also objectively describe facial behaviors of the full face, the errors of automatically detected AU intensities may limit their ability in inferring depression status. Meanwhile, gaze and head pose can only reflect some specific facial behaviors and fail to include depression-related behaviors occurred in other local facial regions. 

\textbf{Graph-based facial landmarks} Table \ref{table:3} compares the depression recognition results achieved by applying the explored CNN and GNN to process clip-level facial landmark representations. The results show that the explored GNN clearly outperforms the CNN, suggesting that facial landmark time-series is more suitable to be represented as a graph. Moreover, we show that our fusion module is able to combine the latent representations produced from CNNs and GNN, as the best results is achieved by the CNN-GNN model.

\begin{figure}[ht]
    \centering
    \includegraphics[width=0.5\textwidth]{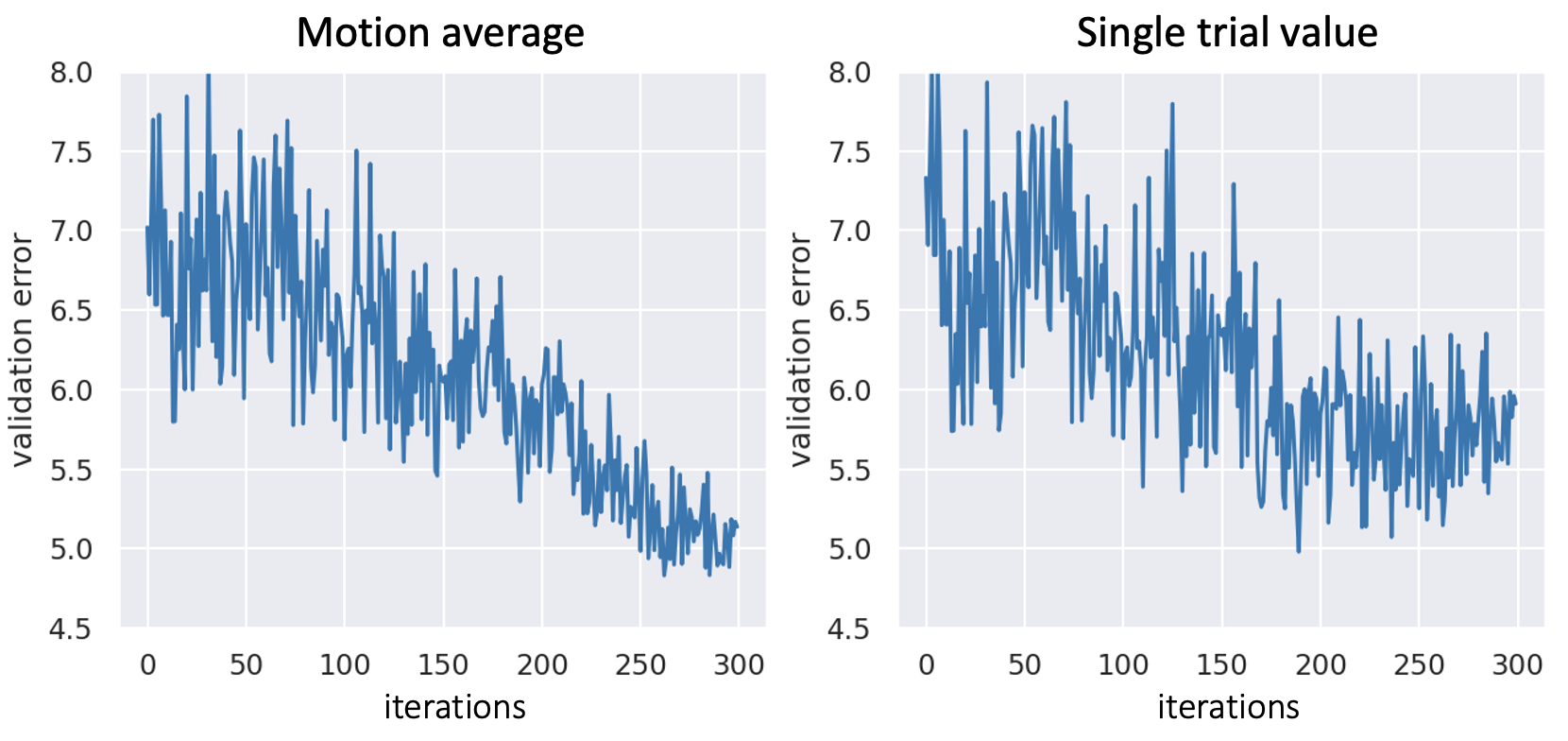}
    \caption{Comparison of the validation error for GraphNAS using motion average (left) and value of a single trial (right)}
    \label{fig:4}
\end{figure}

\textbf{Motion average loss:} Finally, we also evaluate the proposed motion average loss in Fig \ref{fig:4}. As we can see, Fig \ref{fig:4}, we demonstrate the learning curve of the controller for each time-step, i.e., the average validation error of the architectures sampled at each step. When adopting our motion average loss, the validation loss curve has decreased variance and less fluctuation. When a single validation error is adopted as the reward signal, the variance is consistently larger and eventually converges to a sub-optimal policy compared to the policy obtained by using our motion average loss.

\begin{table}\centering
    \begin{tabular}{ p{2cm} p{1.9cm} cc}
     \hline
     \hline
       & Architecture & RMSE  & MAE    \\
      \hline


     \multirow{2}{*}{Single-modal} & CNN&   5.55 &  4.50  \\
                    & GNN   & 4.93 & 3.82 \\
    \hline
    \multirow{2}{*}{Fusion} & CNN&   5.07 &  4.11  \\
                    & GNN   & 3.96 & 3.23 \\
     \hline
     \hline

    \end{tabular}
    \caption{Results achieved by  different facial landmark feature extractor settings.}
    \label{table:3}
 \end{table}

\section{Conclusion}

\noindent In this paper, we propose the first Neural Architecture Search approach to jointly explore optimal CNN-GCN feature extractors and a fusion module for predicting depression from multiple facial attributes. The results show that the model explored by our approach can learn superior depression-related features from all facial attributes, and the fusion module can further enhance the performance by combining depression-related supplementary cues from all facial attributes, with extremely large improvements over the existing state-of-the-art (30\% RMSE improvements). In summary, our study provides solid evidences and a strong baseline for applying NAS to automatic depression analysis.

\bibliographystyle{IEEEtran}
\bibliography{IEEEabrv,reference}

\end{document}